\documentclass{article}

\PassOptionsToPackage{numbers, compress}{natbib}

 \usepackage[dblblindworkshop, final]{neurips_2025}

\workshoptitle{UrbanAI: Harnessing Artificial Intelligence for Smart Cities}

\usepackage[utf8]{inputenc} 
\usepackage[T1]{fontenc}    
\usepackage{hyperref}       
\usepackage{url}            
\usepackage{booktabs}       
\usepackage{amsfonts}       
\usepackage{nicefrac}       
\usepackage{microtype}      
\usepackage{xcolor}         
\usepackage{amsmath}
\usepackage{subcaption}
\usepackage{graphicx} 
\usepackage{placeins}
\usepackage{float}    

\title{Recovering Origin–Destination Flows from Bus CCTV: Early Results from Nairobi and Kigali}

\author{
  Nthenya Kyatha \\
  University of Massachusetts Amherst \\
  Amherst, MA, USA \\
  \texttt{mkyatha@umass.edu}
  \And
  Jay Taneja \\
  University of Massachusetts Amherst \\
  Amherst, MA, USA \\
  \texttt{jtaneja@umass.edu }
}

\begin{document}

\maketitle



\begin{abstract}
Public transport in sub-Saharan Africa (SSA) often operates in overcrowded conditions where existing automated systems fail to capture reliable passenger flow data. Leveraging onboard CCTV already deployed for security, we present a baseline pipeline that combines YOLOv12 detection, BotSORT tracking, OSNet embeddings, OCR-based timestamping, and telematics-based stop classification to recover bus origin--destination (OD) flows. On annotated CCTV segments from Nairobi and Kigali buses, the system attains high counting accuracy under low-density, well-lit conditions (recall $\approx$95\%, precision $\approx$91\%, F1 $\approx$93\%). It produces OD matrices that closely match manual tallies. Under realistic stressors such as overcrowding, color-to-monochrome shifts, posture variation, and non-standard door use, performance degrades sharply (e.g., $\sim$40\% undercount in peak-hour boarding and a $\sim$17 percentage-point drop in recall for monochrome segments), revealing deployment-specific failure modes and motivating more robust, deployment-focused Re-ID methods for SSA transit.
\end{abstract}

\section{Introduction}
\label{sec:intro}

Public transport in sub-Saharan Africa (SSA) is dominated by buses and matatus that frequently operate well beyond capacity. Reliable Origin--Destination (OD) matrices are critical for route design, scheduling, equity analysis, and assessing the impacts of electric fleet adoption, yet most agencies still rely on costly manual surveys that provide only partial and infrequent insight \cite{zalewski2019survey}.

A range of automated technologies has been explored for OD estimation, including Automated Fare Collection (AFC) systems, Automated Passenger Counters (APCs), RFID-based boarding systems, Bluetooth, and mobile phone data, as well as multi-source approaches combining GIS data with IC card records \cite{jafari2021AFC,liu2021APC,gonzalez2020RFID,ozbay2017BT,larijani2015Mobile,kong2021GISIC}. While effective in high-resource contexts, these methods suffer from high cost, maintenance overhead, limited coverage, and privacy concerns, and are rarely deployed at scale in SSA bus networks. They also typically provide only aggregate boarding and alighting counts, making it difficult to track individuals through the vehicle or distinguish official stops from opportunistic roadside pickups.

Video-based re-identification (Re-ID) offers a complementary path by leveraging existing onboard CCTV. Prior work has combined YOLO detection with Re-ID features for passenger matching \cite{shimada2019YOLO}, or focused on head-only features and pose-guided attention \cite{zhao2022Head,miao2019Pose}, but these approaches are brittle under occlusion, posture changes, and clothing similarity. TransitReID \cite{huang2025transitreid} introduces occlusion-resistant embeddings and hierarchical dynamic matching tailored to transit settings, yet it assumes relatively controlled, higher-resolution, predominantly color footage and does not account for illegal roadside stops or frequent color-to-monochrome switching. As a result, it has not been evaluated in the low-resolution, modality-shifting, overcrowded CCTV environments typical of SSA buses.

These gaps motivate our baseline pipeline for OD inference in SSA buses. Our design leverages existing CCTV and telematics streams without additional hardware, combining passenger detection and tracking, OCR-based timestamping, Region-of-Interest (ROI) based event counting at doors, and cross-camera association with stop classification. This four-stage architecture (detailed in Sec.~\ref{sec:system}) produces stop-level OD matrices while surfacing deployment-specific challenges. In contrast to prior work, we deploy the system directly on Nairobi and Kigali buses, demonstrating feasibility under real-world conditions and highlighting challenges unique to SSA transit, such as extreme overcrowding, non-standard door usage, frequent color-to-monochrome CCTV transitions, and illegal roadside stops that remain underexplored in existing literature.

\section{System Design}
\label{sec:system}

Our baseline system integrates four stages (Fig.~\ref{fig:pipeline} in Appendix~\ref{app:extra-figs}): 
(i) per-camera passenger detection and local tracking using YOLOv12 with BotSORT and Re-ID features; 
(ii) OCR-based timestamp extraction, combined with the known frame rate (FPS) to align events at one-second resolution; 
(iii) Region-of-Interest (ROI) based IN/OUT counting at each door; and 
(iv) cross-camera association and telematics alignment, which link front- and exit-camera tracklets where necessary, classify stop locations, and ultimately construct OD matrices.

\subsection{Detection and local tracking}

For both Cam-A and Cam-B, passengers are detected in every frame using YOLOv12, chosen for its balance of accuracy and speed on crowded, low-resolution footage. BotSORT maintains temporal consistency by assigning local IDs to each detection, augmented with appearance features from a lightweight Re-ID backbone (OSNet). This produces per-camera tracklets that capture each passenger's short-term motion through the scene. Importantly, both cameras are treated as bidirectional sensors: each ROI crossing may correspond to either a boarding or an alighting, since real-world behavior in Nairobi and Kigali often diverges from the intended "front-in, rear-out" design. These per-camera tracklets form the basis for both local counts and cross-camera associations.

\subsection{Cross-camera Re-Identification}

Tracklets from Cam-A (front aisle) and Cam-B (exit door) are then associated using appearance embeddings. Let $\mathbf{e}_i$ and $\mathbf{e}_j$ denote the OSNet embeddings for tracklets $i$ and $j$ from Cam-A and Cam-B, respectively. We compute cosine distance
\begin{equation}
d_{ij} = 1 - \frac{\mathbf{e}_i^\top \mathbf{e}_j}{\|\mathbf{e}_i\|_2 \, \|\mathbf{e}_j\|_2}.
\end{equation}
These distances populate a cost matrix $D = [d_{ij}]$ over active tracklets in the two views. We apply the Hungarian algorithm to obtain a minimum-cost bipartite matching, subject to a gating threshold $d_{ij} \le \tau_{\text{reid}}$; pairs that exceed the threshold remain unmatched and start new global trajectories. This cross-camera Re-ID step links a passenger's trajectory across the two streams, enabling reconstruction of complete journeys from boarding at the entrance to alighting at the exit and transitioning from per-camera IDs to global passenger trajectories needed for OD matrix construction.

\subsection{OCR Timestamp Extraction and Aggregation}

Each CCTV frame contains a time overlay produced by the recorder. To synchronize vision-based events with telematics data, we apply OCR to extract these timestamps. Consecutive frames often share the same overlay value, so we combine the OCR strings with the known frame rate (FPS) to generate a reliable per-second timeline. Boarding and alighting events detected from ROI crossings are aggregated into per-second IN/OUT counts, producing a time-stamped event log that aligns with bus stop arrivals.

\subsection{Baseline ROI counting}

In the baseline configuration, ROI-based counting at each door treats every tracked box that crosses the ROI as a candidate boarding or alighting event, independent of the door state. This simple strategy works well when passenger density is low and few people linger near the doors, but it produces false positives when riders stand or move near the exit while the bus is in motion, or when conductors repeatedly enter and leave the ROI without actually alighting. These errors are most pronounced at the exit door (Cam-B), where passengers frequently cluster around the steps.

\subsection{Telematics Integration}

The telematics stream provides vehicle data such as GPS coordinates, wheel-based speed, odometer reading, voltage, current, and various vehicle states. In this work, we primarily use wheel-based speed, odometer reading, and GPS to estimate distance along the route and identify potential stop periods. In the simplest baseline formulation, candidate stops are detected whenever the bus comes to a complete halt (wheel speed $v_t = 0$) and the GPS position lies near a known stop coordinate.

Let $v_t$ denote wheel-based speed at time $t$, $g_t$ the GPS coordinate, and $\mathcal{S}$ the set of official stop coordinates. We define the distance to the nearest stop as
\begin{equation}
\operatorname{dist}(g_t, \mathcal{S}) = \min_{s \in \mathcal{S}} \| g_t - s \|_2.
\end{equation}
We label an \emph{official stop} if $v_t = 0$ and $\operatorname{dist}(g_t, \mathcal{S}) \le \delta_{\text{gps}}$. To better capture real-world behavior, we extend this baseline with a more flexible notion of \emph{illegal stops}: we label an illegal stop if either (i) $v_t = 0$ and $\operatorname{dist}(g_t, \mathcal{S}) > \delta_{\text{gps}}$, or (ii) $0 < v_t < \tau_{\text{slow}}$ and the door-status signal indicates an open door, coinciding with ROI events. These rules capture passengers boarding or alighting amidst congestion and at unscheduled roadside points. By combining ROI-based events with this stop classification, the system distinguishes official passenger exchanges at bus stops from opportunistic exchanges during slow traffic or at off-stop locations.

\subsection{OD matrix construction}

Finally, cross-camera matches, per-second event logs, and classified stops are combined. For each reconstructed passenger trajectory, the system identifies a boarding stop and an alighting stop, incrementing the corresponding cell in the OD matrix. This process yields detailed flows between official stops, while also capturing demand at illegal roadside pickups and drop-offs. The OD matrices produced in this way serve as the core output of the baseline system, supporting equity analysis and transit planning.



\subsection{Hybrid detection with head-only fallback}
\label{sec:hybrid_det}

In practice, the baseline detector struggles most under extreme crowding at the exit door, where only passenger heads are visible inside the ROI. To mitigate this, we introduce a hybrid detection strategy that uses a dedicated head detector when local density is high. Concretely, we first run the full-body YOLOv12 model and count the number of tracked boxes whose centers fall inside the door ROI. If this number exceeds a threshold of five people, we treat the scene as overcrowded and switch to a head-only detector trained on CrowdHuman-style annotations. In these frames, IN/OUT events are derived from the head detections rather than the full-body boxes. When the number of people inside the ROI falls back below the threshold, the system returns to full-body detection. This hybrid scheme preserves the simplicity of the baseline while providing a fallback that is better aligned with the visible signal under peak loads.

\subsection{Door-state aware counting}
\label{sec:door_state}

We also refine ROI-based counting by conditioning it explicitly on door status. In this improved configuration, a door-status signal derived from vehicle telemetry and/or visual cues is used to filter candidate events: ROI crossings at the exit door (Cam-B) are only logged when the door is physically open. This prevents false counts from passengers who stand or move near the exit while the bus is in motion, and from transient occluders such as conductors leaning into the ROI. As we show in Sec.~\ref{sec:findings}, this door-state aware counting substantially improves exit-door accuracy and reduces counting error without changing the underlying detector or tracker.

\section{Findings So Far}
\label{sec:findings}


We evaluate the baseline and its variants on 11 manually annotated CCTV segments from Nairobi and Kigali buses, each 3- 8 minutes long, covering a mix of low-density, peak-hour, colour, and monochrome scenes at both doors. For each segment, we annotate per-door entry and exit counts and derive OD matrices by pairing boarding and alighting stops. We report accuracy (fraction of correctly counted passengers), mean absolute error (MAE, in passengers), and the number of complete misses (segments where a method records zero events while the ground truth is non-zero).

\begin{table}[t]
  \centering
  \caption{Baseline pipeline accuracy under different operating conditions.}
  \label{tab:conditions}
  \begin{tabular}{lccc}
    \toprule
    Condition & Detection & Tracking & OD Matrix Accuracy \\
    \midrule
    Ideal (low density, good lighting) 
      & 40/42 correct, +4 FP 
      & Stable IDs 
      & Matches manual tallies \\
    Overcrowding (peak hour) 
      & Missed detections 
      & Frequent ID switches 
      & $\sim$40\% undercount \\
    Poor lighting / shadows 
      & Degraded confidence 
      & Unstable IDs 
      & Diverges at exits \\
    Posture changes 
      & Robust detections 
      & Fragmented tracklets 
      & Incomplete trips \\
    Color $\leftrightarrow$ B/W switching 
      & Robust 
      & Inconsistent features 
      & $\sim$17\% degradation \\
    Non-standard door use 
      & Robust 
      & Ambiguous direction 
      & Noisy OD matrix \\
    \bottomrule
  \end{tabular}
\end{table}

\begin{figure*}[t]
  \centering
  \includegraphics[width=\linewidth]{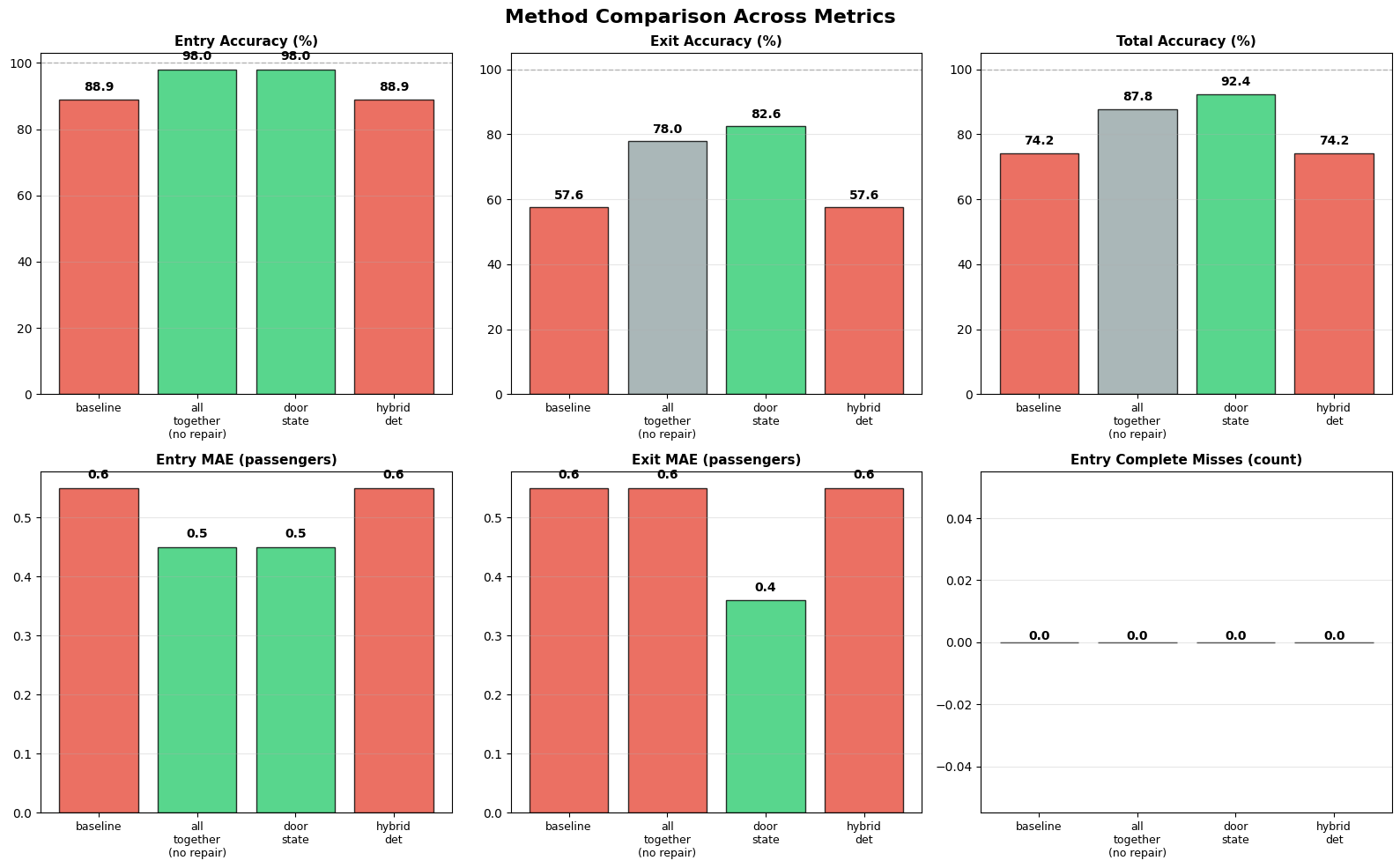}
  \caption{Method comparison across 11 annotated segments. 
  We report entry, exit, and total accuracy (top row) and mean absolute error (MAE, in passengers; bottom row) for the baseline pipeline, the door-state aware variant (\texttt{door\_state}), the hybrid head-only detector configuration (\texttt{hybrid\_det}), and an \texttt{all\_together (no id repair)} variant that combines improved components without identity repair. 
  Entry performance is high for all methods, while exit accuracy and exit MAE improve substantially when incorporating door-state gating and related telematics cues.}
  \label{fig:method-comparison}
\end{figure*}

Across all segments, the baseline system achieves high entry performance but struggles at the exit door. Aggregated over the 11 videos, entry accuracy is $88.9\%$ with entry MAE of $0.6$ passengers, while exit accuracy drops to $57.6\%$ with exit MAE of $0.6$ passengers (Fig.~\ref{fig:method-comparison}). Total accuracy over both doors is $74.2\%$. These aggregate results are consistent with our earlier condition-wise analysis(summarised in Table ~\ref{tab:conditions}: in low-density, well-lit footage the baseline counts 40 of 42 passengers correctly (recall $\approx 95\%$) with 4 false positives, and OD matrices match manual tallies, whereas in overcrowded peak-hour boarding at the exit door we observe undercounts of roughly $40\%$, and in monochrome footage we see an average degradation of about $17\%$ points in recall relative to comparable color segments. Posture variation and non-standard door usage further fragment tracklets and introduce ambiguity in directionality, leading to incomplete trip reconstruction even when detection and timestamping remain robust. More details on the data are in Appendix \ref{app:data}

We then compare three variants to the baseline: (i) \texttt{hybrid\_det}, which introduces the head-only detector fallback when more than five people occupy the exit-door ROI; (ii) \texttt{door\_state}, which gates ROI crossings by the door-status signal; and (iii) \texttt{all\_together (no id repair)}, which combines the improved components without applying cross-camera identity repair. Entry counts are already near ceiling, and all methods achieve similar entry performance: both \texttt{door\_state} and \texttt{all\_together} reach $98.0\%$ entry accuracy with MAE $0.5$ passengers, while the baseline and \texttt{hybrid\_det} remain at $88.9\%$ with MAE $0.6$. The main gains appear at the exit door: compared to the baseline's $57.6\%$ exit accuracy, \texttt{all\_together} improves to $78.0\%$ and the door-state aware variant further to $82.6\%$, reducing exit MAE from $0.6$ to $0.4$ passengers. In contrast, the \texttt{hybrid\_det} configuration does not improve exit accuracy over the baseline on these segments. Overall, total accuracy rises from $74.2\%$ (baseline) to $87.8\%$ for \texttt{all\_together} and $92.4\%$ for the door-state aware configuration. Per-video results are presented in Appendix~\ref{app:per_clip_results}.

These results suggest that simple, domain-informed cues, especially door-state gating of ROI events and more flexible stop classification that accounts for near-zero motion with open doors, can substantially mitigate the worst failure modes of the baseline at the exit door, even without introducing more complex Re-ID architectures or identity-repair modules. At the same time, the remaining exit errors under extreme crowding and modality shifts indicate that more robust embeddings and trajectory-aware repair will be necessary for reliable OD inference at scale.

\section{Challenges and Discussion}
\label{sec:discussion}

The above findings reveal several deployment-specific challenges in sub-Saharan bus environments. \textbf{Overcrowding and occlusion} at doorways frequently obscure passengers, fragmenting tracklets and driving undercounts of up to $40\%$. \textbf{Posture changes} (standing, sitting, leaning) disrupt appearance-based embeddings, leading to broken Re-ID continuity. \textbf{Modality shifts}, where cameras switch between color and monochrome, destabilize feature embeddings and degrade OD accuracy by roughly $17\%$. \textbf{Visual similarity} among riders in school uniforms or similar clothing causes identity switches across tracklets. Finally, \textbf{non-standard door usage}, with boarding and alighting at both front and exit doors, introduces ambiguity into ROI-based event classification.

Our ablations indicate that lightweight, domain-informed cues can substantially reduce these errors. Door-state-aware counting and a more flexible stop definition that includes near-zero motion with open doors improve exit accuracy from $57.6\%$ (baseline) to $82.6\%$ and total accuracy from $74.2\%$ to $92.4\%$, without changing the detector or Re-ID backbone. In contrast, a hybrid head-only detector provides limited benefit on our current clips, suggesting that better use of temporal and door context is more valuable than simply adding another detector. Nevertheless, remaining failures under extreme crowding and modality shifts indicate that more robust embeddings and trajectory-aware identity repair will be needed for reliable OD inference at scale, and that deploying methods such as TransitReID in this low-resolution, color/monochrome, roadside-stop setting remains an open challenge.

\textbf{Data sharing and reproducibility.}
Work with in-vehicle CCTV necessarily involves agreements with operating companies and start-ups that prioritize passenger safety and privacy. In our case, these agreements prohibit releasing raw video or models that could be repurposed for surveillance beyond the scope of this study. Similar constraints apply to other recent transit Re-ID efforts, which often operate on non-public CCTV datasets. As a result, we currently make public only aggregate OD matrices, per-condition metrics, and implementation details of our baseline pipeline. In future work, we plan to explore privacy-preserving benchmarks (e.g., synthetic data or heavily anonymised clips) that still allow meaningful comparisons with this baseline.



\section{Conclusion}
\label{sec:conclusion}

We presented a baseline CCTV pipeline for passenger detection, tracking, OCR timestamping, and ROI-based counting fused with telematics data in SSA buses. Using 3--8 minute annotated clips from Nairobi and Kigali, we showed that the system is feasible under light to moderate passenger loads, achieving high entry accuracy and OD matrices that match manual tallies, but is hindered by overcrowding, occlusion, posture changes, modality shifts, and non-standard door usage. By treating both doors as bidirectional and incorporating door-state aware counting and flexible stop definitions, we align the design more closely with real-world operations and substantially improve exit-door accuracy. These early results highlight both the promise and the limitations of CCTV-based OD inference in SSA transit, and point toward the need for more robust, deployment-focused Re-ID methods.


\section{Next Steps}
\label{sec:next}

To address the remaining failure modes, we plan several extensions. First, we will develop a stronger cross-camera cascade with motion-informed gating and cost-optimal assignment to reduce mismatches under crowding and non-standard door usage. Second, we will incorporate dual-branch embeddings (head and body crops) with grayscale normalization to better handle partial visibility and color/monochrome shifts. Third, we will conduct controlled robustness studies that vary posture and crowd density to quantify their impact on ID stability. Finally, we will extend deployments to longer time windows, enabling analysis of equity impacts, for example, identifying which neighborhoods benefit most from electrified routes and which depend heavily on unscheduled roadside stops.

\bibliographystyle{abbrvnat}
\small \bibliography{ref.bib}
\appendix

\section{Supplementary Material}

We include data and annotation details, the full baseline pipeline diagram, and per-clip counting results to support reproducibility and interpretation of the main findings.

\subsection{Data and annotation details}
\label{app:data}

In both cities, bus configurations differ by door count, camera count and placement. \textbf{Kigali} buses generally have two doors, an entrance at the front and an exit in the middle of the cabin. Cameras: Driver Cabin ($Cam\text{-}Dr_{k}$), placed directly above the driver's cabin and capturing only the driver; Dashboard ($Cam\text{-}Dash_{k}$), facing outward towards oncoming traffic; Front Door ($Cam\text{-}A_{k}$) positioned at the front of the bus and faces down the aisle; captures boardings, front-door alightings and initial movements on the aisle; Exit Door ($Cam\text{-}B_{k}$) located opposite the exit door, facing the doorway; captures alightings and occasional rear-door boardings; and where present Rear ($Cam\text{-}Re_{k}$) placed at the rear and facing forward for extended aisle trajectories. In contrast, \textbf{Nairobi} buses have a single front door that serves as both the entrance and the exit. Like the buses in Kigali, these buses are equipped with Driver Cabin Cameras ($Cam\text{-}Dr_{n}$) and Dashboard Cameras ($Cam\text{-}Dash_{n}$). Additionally, each bus has a Front Door Camera ($Cam\text{-}A_{n}$) located near the front facing the door, capturing both boarding and alighting. 

\subsubsection{Recording \& telematics.} In both cities, CCTV footage runs from $\sim$05{:}00 to $\sim$23{:}00. Videos are $\approx$30–45 minutes long with on-frame timestamps. Telematics provide GPS (latitude, longitude) and vehicle signals (wheel-based speed, odometer, state-of-charge (SoC), voltage/current, and energy consumed/recuperated/idle/charged/used), each time-stamped with a unique telematics ID.

\subsubsection{Clip selection for tuning/validation.} From \(Cam\text{-}A_{k}\), \(Cam\text{-}B_{k}\), and \(Cam\text{-}A_{n}\), we extract 3–4 minute segments per interval under the following criteria:
(i) uncrowded, (ii) medium-high crowding, (iii) partial occlusions from bus hardware, (iv) color-only or monochrome-only, and (v) alternating monochrome$\leftrightarrow$color (e.g., IR night vision).
For Kigali, paired \(Cam\text{-}A_k\)/\(Cam\text{-}B_k\) views ensure temporal continuity for cross-view ReID. Two highly characteristic videos are densely annotated using Computer Vision Annotation Tool (CVAT) for held-out testing; the final pipeline is evaluated on footage from both cities. Images extracted from the video form part of our Indoor data that will be used to fine-tune the ReID model.

\subsubsection{Data splits.} Because routes repeat daily as terminal-to-terminal \emph{round trips}, we use \emph{day-level, temporally disjoint} splits to prevent temporal/scene leakage while preserving deployment realism. For \textbf{Kigali}, we use the E-bus launch days \textit{March 21} (15{:}00–22{:}00) and \textit{March 26} (04{:}00–22{:}00) for training/validation, and \textit{March 31} (04{:}00–22{:}00) for testing. 

When \textbf{Nairobi} data is included, we mirror this policy with day-level disjoint train+val and test days.Trip/route segmentation uses wheel speed and GPS to partition days into round trips and per-stop dwell intervals; there is no overlap in timestamps, telematics IDs, or trip IDs across splits. We stratify evaluation by time-of-day bins aligned to demand: AM peak (07{:}00–09{:}00), Midday (11{:}00–12{:}00), PM peak (17{:}00–19{:}00), and Off-peak (14{:}00–16{:}00), as well as by lighting (day/night), stream (color/monochrome), and door-ROI occupancy bins
$[0,5),[5,+)$ persons/frame (centers within $\mathcal{P}_c$ or a thin band around $\ell_c$).



\subsection{System Overview}
\label{app:extra-figs}

\FloatBarrier  

\begin{figure}[H]
  \centering
  \includegraphics[width=\linewidth]{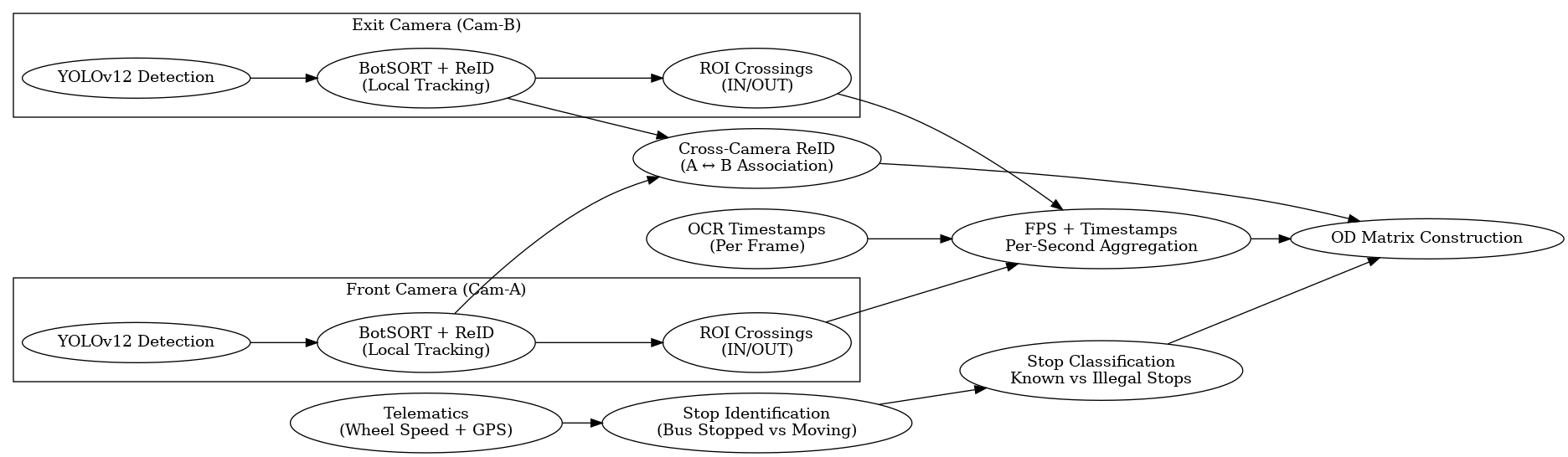}
  \caption{Baseline two-stream pipeline for OD inference. 
  Each camera (Cam-A at the front, Cam-B at the exit) runs YOLOv12 detection with BotSORT+OSNet tracking to produce local tracklets and ROI-based IN/OUT events. 
  OCR-extracted timestamps and FPS align events to a per-second timeline, while telematics (wheel speed and GPS) identify official and illegal stops. 
  These signals are fused to construct stop-level OD matrices; later sections add hybrid head-only detection and door-state aware counting on top of this baseline.}
  \label{fig:pipeline}
\end{figure}
\FloatBarrier

\begin{figure}[H]
    \centering
    \includegraphics[width=\textwidth]{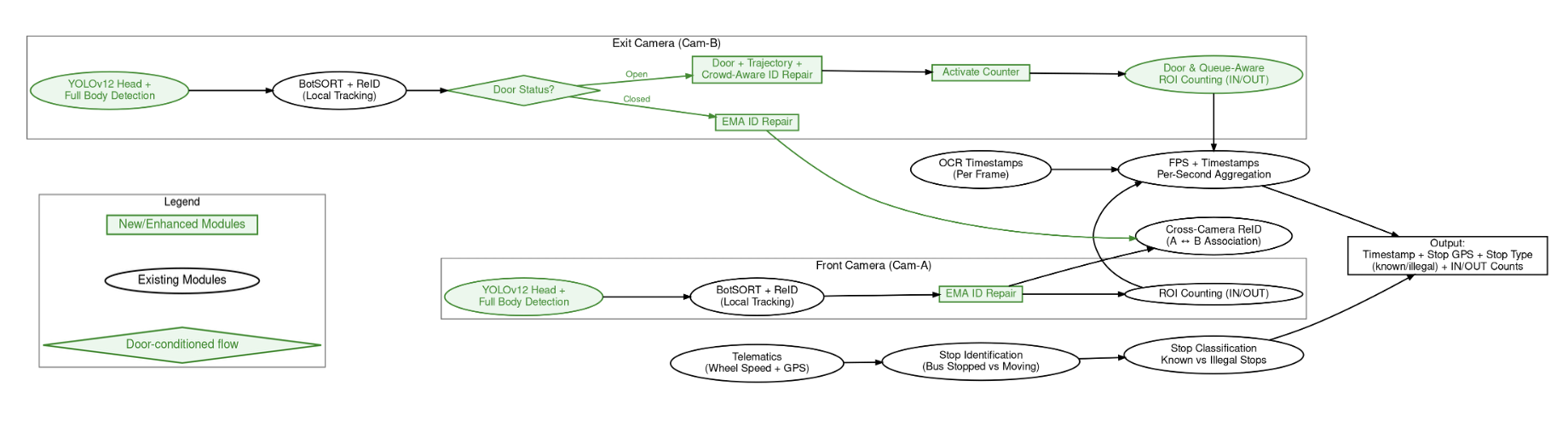}
    \caption{Enhanced Two–stream pipeline with new and enhanced modules highlighted in green. 
    Front camera (Cam A): YOLOv12 Full body with head detection as a fallback in crowded scenes for detection $\rightarrow$ BoT–SORT with re-identification (local tracking) $\rightarrow$ \textbf{EMA identity repair} $\rightarrow$ \textbf{door and queue aware ROI counting}. 
    Exit camera (Cam B): YOLOv12 Full body with head detection $\rightarrow$ BoT–SORT with re–identification $\rightarrow$ \textbf{door plus trajectory plus crowd aware identity repair} (gated by the \textbf{door state} open or closed signal) $\rightarrow$ \textbf{door and queue aware ROI counting}. 
    Shared modules: OCR timestamps per frame and frame rate aggregation for per second tallies, cross camera re–identification (A$\leftrightarrow$B association), telematics signals for stop identification and stop classification (known vs illegal), and final origin–destination matrix construction. 
    This figure summarizes the full system but is not central to the experimental comparisons in the main text.}
    \label{fig:two-stream-supp}
\end{figure}

\subsection*{Planned direction: making counting aware of passenger queues}
Passengers often queue immediately outside the door and are visible in the exit camera before entry. In current footage these queued passengers can be intermittently detected and occasionally counted if the system does not have explicit knowledge of the queue. The next development step is to \emph{streamline the counter to be queue aware} so that people waiting outside are not counted until they truly cross into or out of the vehicle.

\paragraph{Queue–aware counter design goals.}
\begin{itemize}
  \item \textbf{Explicit queue region outside the door.} Add an exterior polygon adjacent to the door and require a valid IN count to include a transition from the exterior polygon to the interior line of interest within a short temporal window.
  \item \textbf{Door state gating.} Suppress all IN and OUT events while the door is closed and for a short grace period after closure to avoid late detections.
  \item \textbf{Direction by distance to line.} Use change in signed distance to the line of interest rather than a fixed axis so that oblique door lines are handled correctly.
  \item \textbf{Temporal hysteresis around the threshold.} Require sustained presence on the interior side for a few frames before finalizing a count, which reduces flicker from partial occlusions at the step.
  \item \textbf{Density–aware policy switch.} When the door region is crowded, enable the trajectory aware repair and stricter queue rules; otherwise use the lighter EMA repair path.
  \item \textbf{Cross camera de–duplication.} Use cross camera re–identification to avoid double counting when a person is briefly visible in both views during boarding or alighting.
\end{itemize}

\noindent These changes align the pipeline with typical bus operations where passengers accumulate in a queue just outside the door. By delaying counts until a verified transition from the queue region into the cabin or from the cabin into the exterior, we expect fewer spurious counts and tighter agreement with manual tallies.

\paragraph{Identity–repair policy.}
In preliminary tests, off-the-shelf \textsc{BoT-SORT \cite{Aharon2022BoTSORT}}+\textsc{ReID \cite{Zhou2019OSNet}} degraded under heavy occlusion near the door, producing frequent identity switches and fragmented trajectories. To address this, we implemented three identity-repair modules layered on top of tracking:

\begin{itemize}
  \item \textbf{EMA Repair (short-occlusion repair):} We employ an EMA-based temporal smoother over per-track embeddings as an engineering choice inspired by EMA usage in ReID for teacher/anchor updates (e.g., mean-teacher and EMA-anchor approaches), rather than following a specific prior that used EMA as a pooling operator\cite{Ge2020MMT,Hou2019VRSTC, Li2019GLTR,TarvainenValpola2017MeanTeacher} . Prior video ReID typically aggregates frame features via average or attention pooling. The system maintains per-track embedding histories; when a track goes missing and a new identity appears nearby within a brief window, we stitch them if spatial gates (intersection-over-union, or normalized center distance) and appearance similarity ($1-\cos(\cdot)$) pass. A small temporal pooler (exponential moving average or a lightweight long short-term memory style aggregator) summarizes recent embeddings.
  \item \textbf{Door-Aware Repair:} In this case, we build on EMA repair by adding a layer where we consider the status of the door which in some cases leads to variation in the quality of the video and also occlusion with passengers coming in or leaving, or crowding at the door as they make their way to the seats. We dynamically adjusts appearance and spatial gates based on door state and geometry overlap, for example both boxes inside the door region of interest or crossing the boundary, with ambiguity and ownership guards to avoid risky merges.
  \item \textbf{Door-Aware + Trajectory Repair:} We observe that  Door Aware repair performs worse due to occlusion between passengers, movement of passengers inside the bus, most of them wearing similar colored clothes, and sudden change in video quality from color to black and white which affects the reidentification model leading to high rate wrong id assignement. We add another layer on the Door aware repair. This layer adds velocity-smoothed motion prediction and spatial conflict checks using a grid index and an exclusion intersection-over-union (IoU) test to favor physically plausible reconnections during dense flows.
\end{itemize}

This preliminary exploration tuned \textsc{BoT-SORT} association thresholds and compared the three repair policies, \textbf{EMA Repair}, \textbf{Door-Aware Repair}, and \textbf{Door-Aware + Trajectory}, on two annotated exit door clips using Multiple Object Tracking (MOT) style evaluation. Our expectation was that, in crowded door regions with frequent short occlusions, the door-conditioned methods would outperform the generic EMA repair, while for lighter crowds or aisle-like motion near the door, the cheaper EMA repair would suffice.

\begin{figure}[H]
    \centering
    \includegraphics[width=\textwidth]{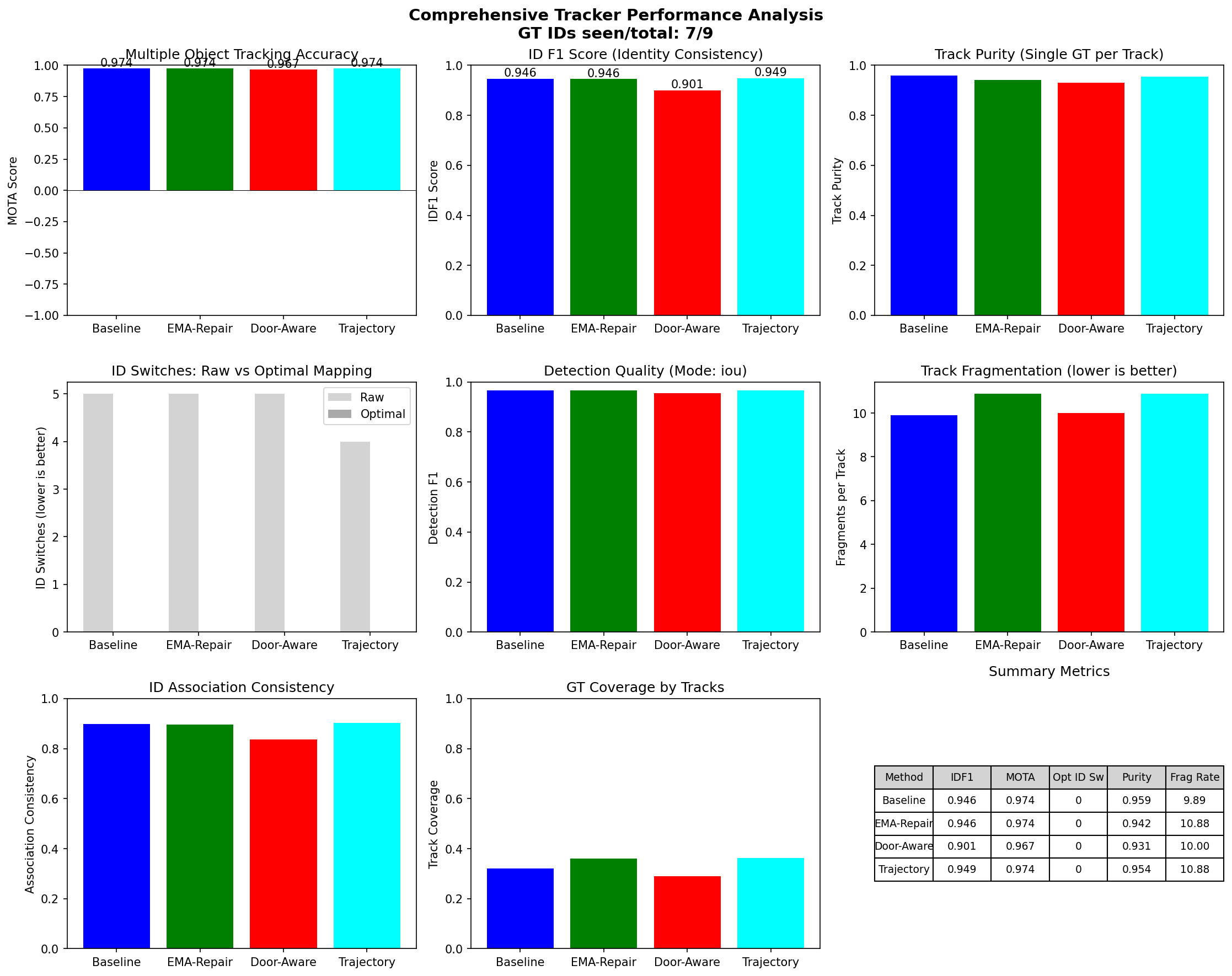}
    \caption{Tracking summary plots from two representative runs. Each panel reports identity consistency (IDF1), overall tracking accuracy (HOTA or MOTA as available), identity switches, detection F1, fragmentation, association consistency, and ground–truth coverage.}
    \label{fig:tracker-summaries}
\end{figure}

\noindent \textbf{Summary of findings.} 

Early results as presented in Figure~\ref {fig:tracker-summaries}, align with these expectations: \textbf{Door-Aware + Trajectory} yields the fewest door-region identity switches and the highest tracking quality in overcrowded scenarios, whereas \textbf{EMA Repair} is competitive and simpler for non-door aisle motion or lighter scenes. Based on these observations, we will prioritize \textbf{Door-Aware Repair} as the default for exit door tracking and enable the trajectory constraints selectively in high crowd, door open intervals to handle dense, short-occlusion flows without unnecessary overhead. 

It's important to note that the repair methods remain effective for shorter videos(3 to 8 minutes). However, for longer durations of 30-45 minutes, there is a significant decline in performance, characterised by increased ID fragmentation. Additionally, due to the poor quality of the footage, the detection model misses a few detection hence further finetuning is required.


\subsection{Per-clip counting results}
\label{app:per_clip_results}

\begin{figure*}[!ht]
  \centering
  \includegraphics[width=\linewidth]{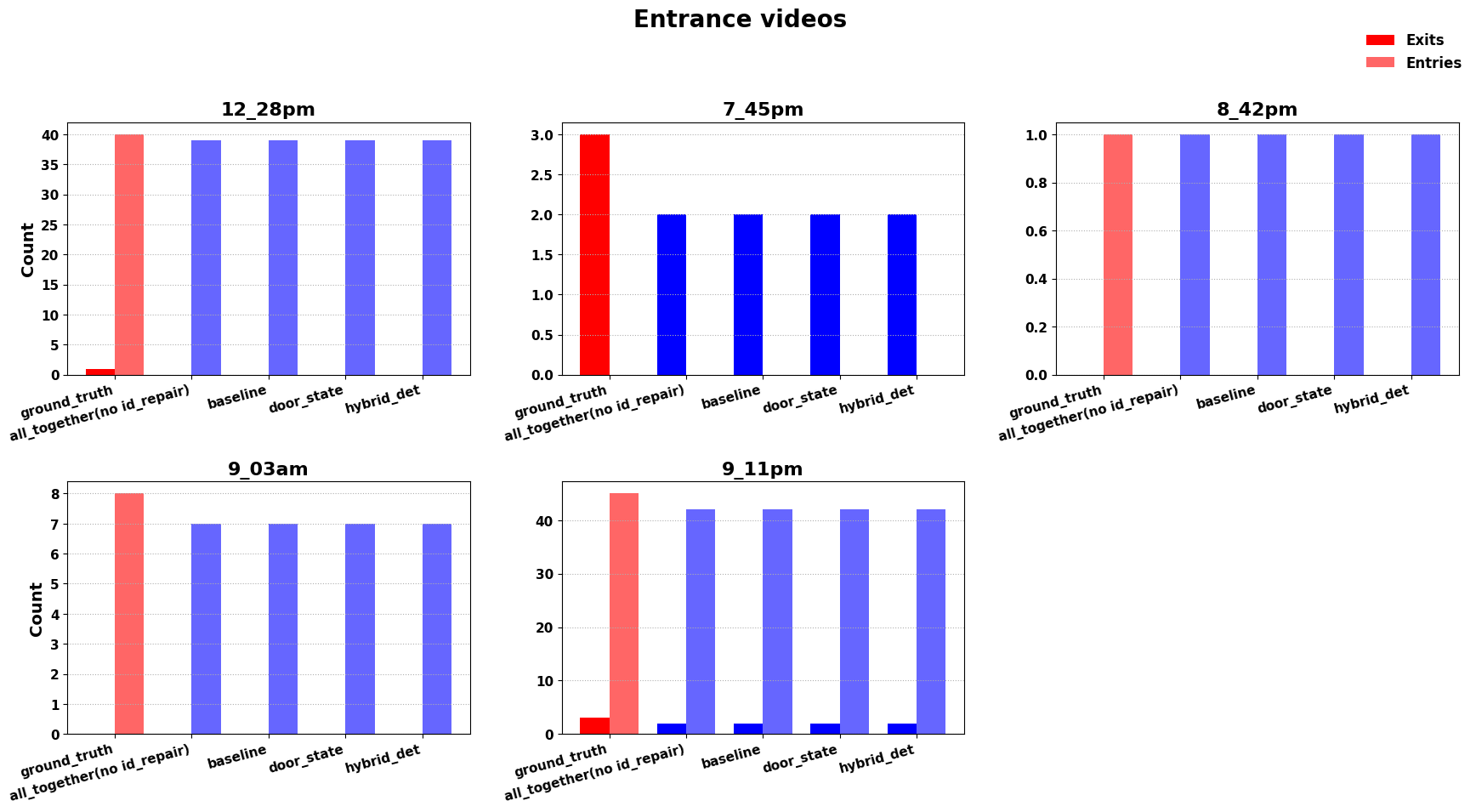}
  \vspace{-0.5em}
  \includegraphics[width=\linewidth]{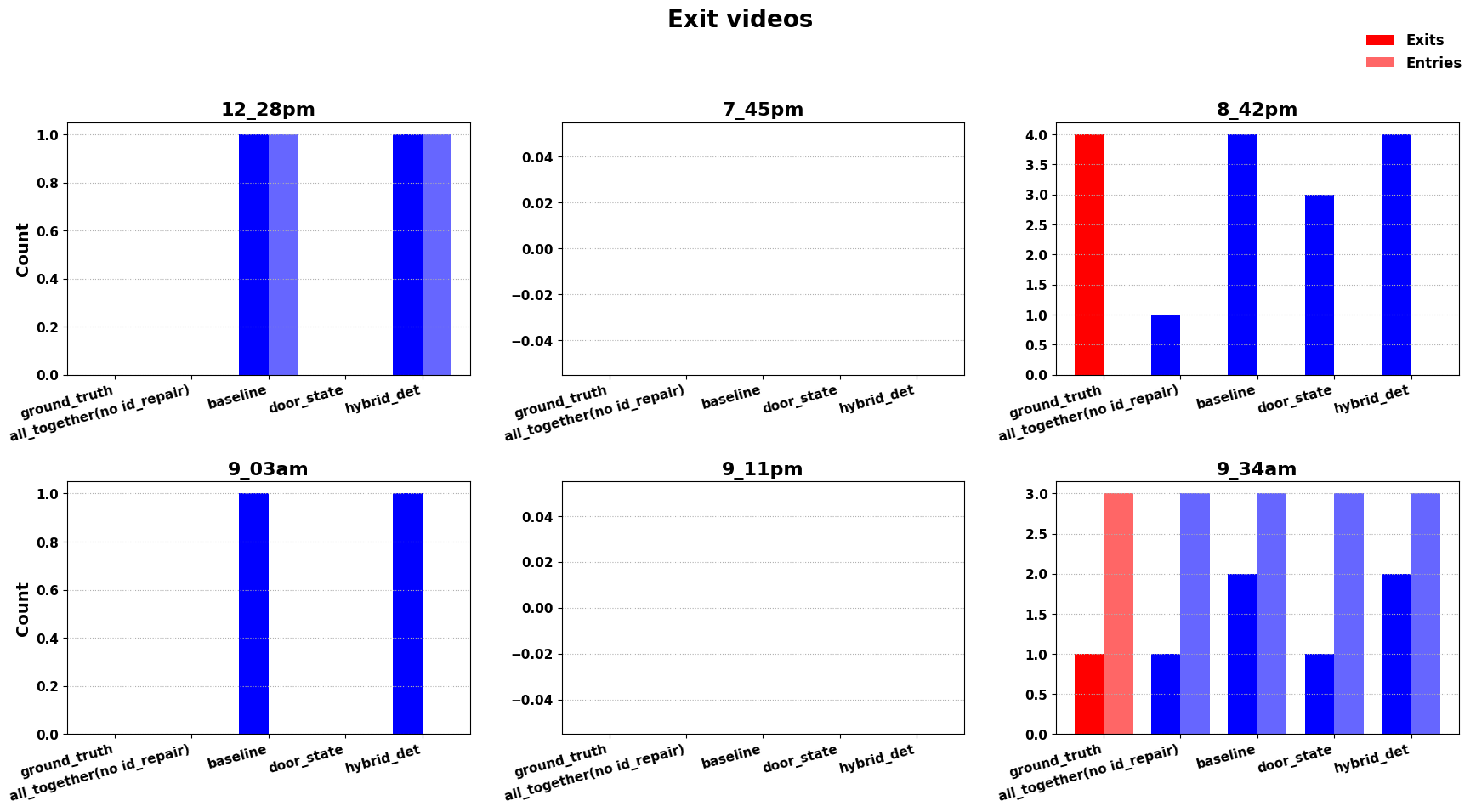}
  \caption{Per-clip entry (top) and exit (bottom) counts across 11 annotated segments. 
  Red bars show ground-truth counts; blue bars show the baseline, \texttt{door\_state}, and \texttt{hybrid\_det} variants. 
  Blank panels (no visible bars) correspond to clips in which no events occurred during the 3--8 minute window, so all methods correctly report zero entries or exits.}
  \label{fig:app-per-clip}
\end{figure*}

\end{document}